\documentclass[10pt,twocolumn,letterpaper]{article}

\usepackage{cvpr}
\usepackage{times}
\usepackage{epsfig}
\usepackage{graphicx}
\usepackage{amsmath}
\usepackage{amssymb}
\usepackage{bm}

\usepackage[breaklinks=true,bookmarks=false]{hyperref}

\cvprfinalcopy 

\def\cvprPaperID{****} 

\begin{document}

\title{Mixture of Counting CNNs: Adaptive Integration of CNNs Specialized to Specific Appearance for Crowd Counting}

\author{Shohei Kumagai$^1$, Kazuhiro Hotta$^1$ and Takio Kurita$^2$\\
$^1$Meijo University\\
{\tt\small 123433009@ccalumni.meijo-u.ac.jp, kazuhotta@meijo-u.ac.jp}\\
 $^2$Hiroshima University\\
 {\tt\small tkurita@hiroshima-u.ac.jp}
}

\maketitle

\begin{abstract}
This paper proposes a crowd counting method. Crowd counting is difficult because of large appearance changes of a target which caused by density and scale changes. Conventional crowd counting methods generally utilize one predictor (\eg regression and multi-class classifier). However, such only one predictor can not count targets with large appearance changes well. In this paper, we propose to predict the number of targets using multiple CNNs specialized to a specific appearance, and those CNNs are adaptively selected according to the appearance of a test image. By integrating the selected CNNs, the proposed method has the robustness to large appearance changes. In experiments, we confirm that the proposed method can count crowd with lower counting error than a CNN and integration of CNNs with fixed weights. Moreover, we confirm that each predictor automatically specialized to a specific appearance.
\end{abstract}

\section{Introduction}
\label{sec:1}

This paper addresses a crowd counting task. An automatic counting is expected to use in many applications (e.g. crowd counting in surveillance cameras\cite{chan2008privacy}, cars counting in aerial images\cite{kembhavi2011vehicle,arteta2014interactive} and particles counting in microscope images\cite{arteta2014interactive,lempitsky2010learning}). However, those objects are manually counted by observers now. Such manual counting can not treat a lot of images. Moreover, counting result becomes subjective. Therefore, an automatically object counting method is really required to obtain objective counting results.

\begin{figure}[t]
\begin{center}
\includegraphics[width=0.85\linewidth] {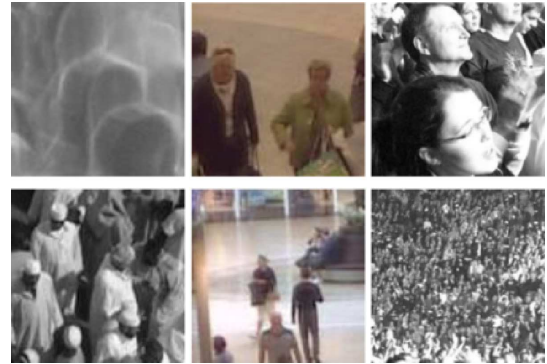}
\end{center}
\caption{Appearance of crowd is largely changed by scale and congestion.}
\label{fig:1}
\end{figure}

However, crowd counting task has three difficult problems in comparison with general image recognition task. The first problem is the occlusion of counting targets. Since counting targets densely exist in an image, the appearance of targets is much different at sparse and dense places. The second problem is that a small target is represented by a few pixels. The last problem is the appearance change caused by scale change of a target. Figure \ref{fig:1} shows the examples of crowd, the appearance of crowd is largely changed by scale change and congestion.

Some counting methods have been proposed to overcome those difficulties. It is reported that regression based methods \cite{chan2008privacy,idrees2013multi,chen2013cumulative,chen2012feature} gave better result than detection based method\cite{felzenszwalb2008discriminatively} for occluded targets in dense regions. Those methods learn the pair of training images and the number of targets contained in the images, and they become robust to occlusion in dense regions. It is reported that multiple features are effective for targets with low resolution \cite{idrees2013multi}. However, the last problem about appearance change caused by scaling has not been sufficiently studied in the approach.

\begin{figure}[t]
\begin{center}
\includegraphics[width=\linewidth] {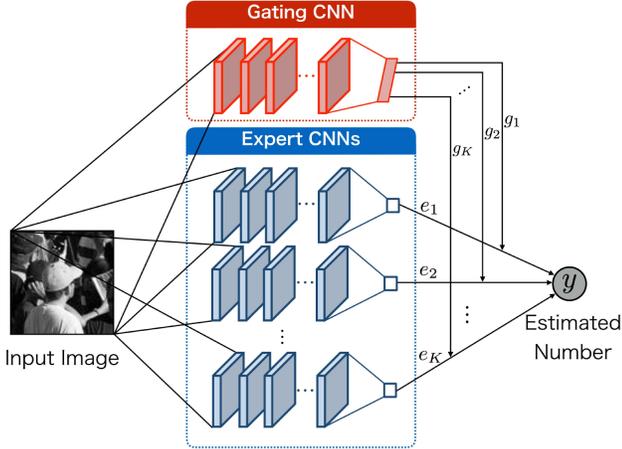}
\end{center}
\caption{The overview of MoC-CNN. Blue square shows expert CNNs. Each expert CNN predicts the number of targets in the input image. Red square shows the gating CNN. To adaptive select expert CNN, gating CNN gives weights for the predictions of expert CNNs.}
\label{fig:2}
\end{figure}

This paper proposes a robust counting method to such appearance changes. As drawbacks of conventional methods, those methods utilize only one predictor (e.g. regression, random forest and CNN) for the various appearances of targets. Such one predictor can not handle the various appearance changes. Thus, we propose multiple predictors specialized to a specific appearance. Those predictors are adaptively selected according to the appearance of targets. By integrating the selected predictors, it is robust to various appearance changes. We use CNN as the predictor because CNN gave the state-of-the-art results in many image recognition tasks \cite{krizhevsky2012imagenet,xu2015discriminative,xiao2015application} in recent years. The effectiveness of CNN for crowd counting is also reported \cite{zhang2015cross}. Our method adaptively integrates some CNNs based on the idea of Mixture of Experts \cite{jacobs1991adaptive} (MoE). Thus, we call our proposed CNN as {\it Mixture of Counting CNNs} (MoC-CNN). The overview of MoC-CNN is shown in Figure \ref{fig:2} .

The MoC-CNN is consists of two types of CNNs. The first CNN specializes to a specific appearance of targets. For example, the specific appearances mean targets in a dense region, targets in a sparse region, small targets and large targets. The specialized CNN is called {\it expert CNN}. We use some expert CNNs specialized to each appearance. The second CNN selects expert CNNs according to the appearance of targets. This CNN is called {\it gating CNN}.

To count targets in an image, the image is fed into both expert CNNs and gating CNN. Each expert CNN predicts the number of targets in the image. On the other hand, gating CNN predicts the probabilities of expert CNNs, and those probabilities are used as the weight for integrating the prediction results by expert CNNs. Thus, the number of target in the image is the weighted mean of results by expert CNNs.

In experiments, we use two crowd count datasets; the UCF Crowd Count (UCF\_CC\_50) dataset and the Mall dataset. We confirm that our method obtain comparable accuracy with the state-of-the-art methods on the both datasets.
In the UCF\_CC\_50 dataset, our method achieved that mean absolute error is 360.8 and mean squared deviation is 488.2. In the Mall dataset, our method achieved that absolute error is 2.75 and mean deviation error is 0.087. In the experiments, we confirm that gating CNN adaptively selects expert CNNs for the appearance of an input image. Moreover, we confirm that each predictor automatically specialized to a specific appearance.

This paper is organized as follows. In Section \ref{sec:2}, we describe related works. The details of the MoC-CNN are described in Section \ref{sec:3}. In Section \ref{sec:4}, we evaluate our proposed method on two crowd counting datasets with various appearances. Finally, we describe conclusion and future works in Section \ref{sec:5}.


\section{Related Works}
\label{sec:2}

Some crowd counting methods have been proposed in recent years. Chan \etal proposed a crowd counting method based on Gaussian process regression\cite{chan2008privacy}. Loy \etal proposed multiple-output ridge regression\cite{chen2012feature}. Chen \etal \cite{chen2013cumulative} proposed a cumulative attribute ridge regression which uses the number of targets as attribute. Those methods can count targets with low counting error. However, they are not essentially robust to scale changes of targets. Therefore, those methods used a perspective map\cite{chan2008privacy} to obtain the similar features from targets of different scales.

Perspective map is very simple and effective for appearance change by perspective. However, this method has two drawbacks. At first, if filming location is changed, we must reset the parameters of the perspective map manually. The second drawback is that the normalization is effective to only targets which are similar sizes (\eg human). However, when the size of targets changes (\eg particles in microscope images and vehicles in areal images), this normalization is not effective.

In other object counting methods, density map is often utilized. The density map was proposed by Lempitsky \etal \cite{lempitsky2010learning}. In general, dot annotations are given to the center of each target. Density map is generated by replacing dots to Gaussian distributions. The advantage of density map is the robustness to vague target on the boundary of image. However, density map does not consider scale changes, those methods do not have sufficient robustness to scale changes. Zhang et al.\cite{pham2015count} proposed density estimation method using CNN. This method also utilizes the perspective map. Thus, this method has the same drawbacks as the methods using perspective map described previously.

In recent years, counting methods using CNN are proposed without the perspective map. Zhang \etal proposed to automatically create a density map considered perspective \cite{zhang2016single}. Moreover, they propose a counting method based on Multi-Column CNN. This CNN has multiple feature extraction units, each feature extraction unit has the filters of different sizes to treat targets with different scales. The features obtained from each CNN are combined into one feature, and the feature is fed into fully-connected layer to predict the number of targets. O\~{n}oro \etal also proposed CNN with multiple feature extraction units \cite{onoro2016towards} to count the without perspective map. In those methods, each feature extraction unit is specialized to an appearance of specific scale (e.g., small, middle and large), those conventional methods can obtain features considering multiple scales. However, those methods use only one predictor. Only one predictor can not handle various appearances of a target, and those methods are not essentially robust to appearance changes.

Therefore, we propose a more robust counting method to appearance changes. Our MoC-CNN integrates expert CNNs, and we adaptively integrate expert CNNs specialized to various appearance in an image. Our method can count targets regardless of the filming location and target size.

\section{Proposed method}
\label{sec:3}

This section describes the details of MoC-CNN. At first, we explain a counting method using expert and gating CNNs in Section \ref{ssec:3-1}. How to train expert CNNs and gating CNN is explained in Section \ref{ssec:3-2}. Finally, the settings of MoC-CNN is explained in Section \ref{ssec:3-3}.

\subsection{Counting by MoC-CNN}
\label{ssec:3-1}

MoC-CNNs predicts the number of targets in an input image using following equation.

 \begin{equation}
 y = \sum_{k=1}^{K}{g}_{k}e_{k},
 \label{eq:1}
\end{equation}
where $K$ is the number of expert CNNs, and $e_k$ is the counting result of the $k$-th expert CNN. $g_k$ is the $k$-th output value of gating CNN, and this is probability value obtained by softmax of output layer. The probability $g_k$ is used as the weight for integrating the expert CNNs. The number of targets $y$ is estimated by weighed sum of output $e_k$ of each expert CNN.

For example, we assume that the 1st expert CNN is specialized to an appearance of dense crowd. If a test image has an appearance of dense crowd, gating CNN increases the weight $g_1$ to 1st expert CNN. Thus, the output of the 1st expert CNN $e_1$ largely reflects to the final counting result $y$. By adaptively selecting, we can obtain the strong robustness than one predictor. Moreover, our MoC-CNN does not require to manually decide the role of each expert CNN, and each expert CNN automatically specialize to a specific appearance by end-to-end manner. The reason is explained in Section \ref{ssec:3-3}.

\subsection{Training MoC-CNN}
\label{ssec:3-2}

This section describes how to train expert CNNs and gating CNN. At first, expert CNN trains using the following loss function.

\begin{equation}
L_{expert} = \frac{1}{N}\sum_{n=1}^{N}(t_n - \sum_{k=1}^{K}{g}_{nk}{e}_{nk})^2,
\label{eq:2}
\end{equation}
where $N$ is mini-batch size, $\sum_{k=1}^{K}{g}_{nk}{e}_{nk}$ is the prediction value for the $n$-th image, and $t_n$ is the ground truth of the $n$-th image. This loss function is the mean squared error between prediction value and ground truth.

MoC-CNN is inspired by MoE. Original MoE can automatically decide that each expert CNN assigns to one of appearance variations from random initial parameters. However, if gating CNN also trains using the loss function (\ref{eq:2}) as well as expert CNN, gating CNN can not train the role of each expert from random parameter, and gating CNN will select only one expert CNN for all images. The possible reason for selecting only one expert is explained follows.

\begin{figure}[t]
\begin{center}
\includegraphics[width=\linewidth]{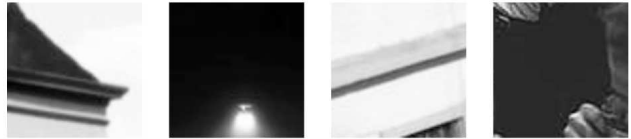}
\end{center}
\caption{Examples of background images with little texture.}
\label{fig:notex}
\end{figure}

At first, training data contain a lot of similar images such as background images with little texture shown in Figure \ref{fig:notex}. At initial stage of training, those similar images are fed into one expert CNN by gating CNN. The expert CNN trains those similar images. On the other hand, few training images are given to other expert CNNs. Therefore, the bias about the number of training images occurs in each expert CNN, and only expert CNN which has many training images obtains high generality. Since gating CNN learns to select expert CNN with low counting error. Gating CNN frequently selects the expert CNN trained by many images, and our network finally uses only one expert CNN. One expert CNN is the same as ordinary CNN, and it is not robust to appearance changes. On the other hand, original MoE integrates neural networks with weak generality. Thus, original MoE can automatically decide the role but expert CNN does not automatically decide the role because CNN has high generality.

To prevent selecting only one expert CNN, gating CNN trains using the following loss function.

\begin{equation}
 L_{gate} = \frac{1}{N}\sum_{n=1}^{N}\{(t_n - \sum_{k=1}^{K}{g}_{nk}{e}_{nk})^2 + \frac{\lambda}{K}\sum_{k=1}^{K}(g_{nk}-\mu_n)^2\},
 \label{eq:3}
\end{equation}
where the first term is the mean squared loss between prediction and ground truth. This is same as the loss function of expert CNN. The second term is variance regularization term. This term works to minimize the variance of outputs of gating CNN, where $\mu_n$ is the mean of output values of gating CNN for the $n$-th training sample. $\lambda$ is trade-off parameter between loss and variance regularization. If gating CNN selects only one expert CNN, the variance becomes large. On the other hand, if gating CNN selects some expert CNNs, the variance becomes small. Therefore, minimization of variance can prevent to select only one expert CNN.

Expert CNNs and Gating CNN are simultaneously trained using each different loss function. The update of output layer in expert and gating CNNs is as follows.

\begin{equation}
\bm{w}_{expert}^k \leftarrow \bm{w}_{expert}^k - \eta_{expert} \frac{\partial L_{expert}}{\partial e_k}\frac{\partial e_k}{\partial \bm{w}_{expert}^k},
\label{eq:4}
\end{equation}

\begin{equation}
\bm{w}_{gate}^k \leftarrow \bm{w}_{gate}^k - \eta_{gate}\frac{\partial L_{gate}}{\partial g_k}\frac{\partial g_k}{\partial \bm{w}_{gate}^k},
\label{eq:5}
\end{equation}
where $\bm{w}_{gate}^k$ is weight vector for the $k$-th output layer of gating CNN, $\bm{w}_{expert}^k$ is weight vector for output layer of the $k$-th expert CNN. $\eta_{expert}$ and $\eta_{gate}$ are learning rate of expert CNN and Gating CNN. Those learning rates are adaptively decided by Adam \cite{kingma2015method}. The parameters on more shallow layers are updated using chain rule as well as the train of ordinary CNN.

{\bf The reason of specialization.} We explain how to automatically specialize expert CNN to a specific appearance. At first, the gradient of loss function in (\ref{eq:4}) is as follows.

\begin{equation}
\frac{\partial L_{expert}}{\partial e_k} = 2g_{k}(\sum_{k=1}^{K}{g}_{k}{e}_{k} - t_{k}),
\label{eq:6}
\end{equation}
where the output of gating CNN $g_k$ works like a learning rate. If the $k$-th expert CNN is selected by gating CNN for a training image, $g_k$ becomes large. Therefore, the $k$-th expert CNN strongly learns the training image．

\subsection{Settings}
\label{ssec:3-3}

\begin{figure}[t]
\begin{center}
\includegraphics[width=\linewidth]{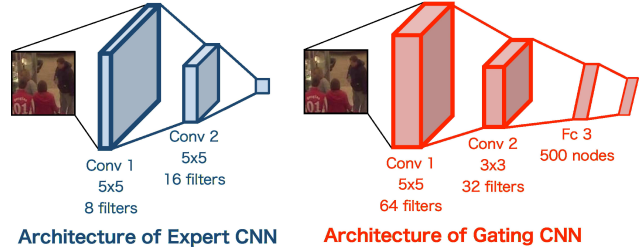}
\end{center}
\caption{Architecture of gating and expert CNNs. Right image is the architecture of expert CNN which is single output network to predict the number of targets. Left image is an architecture of gating CNN which has multiple output layer to predict weights for each expert CNN.}
\label{fig:arch}
\end{figure}

This section describes the details of MoC-CNN setting. At first, the network architecture of expert CNN and gating CNN are shown in Figures \ref{fig:arch}. We prepare $K$ expert CNNs, and the architecture of expert CNN is set to smaller than that of gating CNN. Our method integrates some expert CNNs and assigns the roles to each expert CNN. Therefore, each expert CNN does not learn all training data, and each expert CNN learns only training data given by gating CNN. Thus, expert CNN has small architecture. In this paper, we set the number of experts $K$ to 10 empirically.

{\bf Architecture of expert CNN. }We explain the architecture of Expert CNN. Expert CNN is constructed by 2 convolutional layers, 2 pooling layers and a fully-connected layer. Although pooling layers are omitted in Figure \ref{fig:arch}, we adopt max pooling after each convolution layer. We use max pooling with 2$\times$2 kernel size in the first pooling layer, and the second pooling layer has 3$\times$3 kernel size. We use Batch Normalization\cite{ioffe2015batch} and Exponential Linear Unit(ELU) function \cite{djork2015fast} after each convolutional layer. The architecture is experimentally optimized.

{\bf Architecture of gating CNN. }We explain the architecture of gating CNN. The architecture is also shown in Figure \ref{fig:arch}. Although the output of expert CNNs is single, gating CNN is multi-class classifier. In general, the problem of multiple outputs is more difficult than that of single output. Thus, we consider that gating CNN requires more complex network architecture than expert CNN. The number of filters in each convolution layer of gating CNN is larger than expert CNN. The kernels size in pooling is the same as expert CNN. Batch normalization and ELU are also used. Classification unit of gating CNN consists of two fully-connected layers. We introduce Dropout \cite{hinton2012improving} after the first fully-connected layer, and output layer has softmax function because gating CNN predicts the probabilities for expert CNNs.

{\bf Input image settings. }In our CNN, the size of input image is set to 72$\times$72 pixels. The size refers to \cite{zhang2015cross}. When the size of a test image is larger than 72$\times$72 pixels, we divide the image into grid of 72$\times$72 pixels. If an image is indivisible by 72$\times $72 pixels, image patches below 72$\times$72 pixels are obtained at peripheral region. They are not used for evaluation. The divided patches are fed into MoC-CNN, and we obtain counting results for each patch. Counting results of image patches are summarized, and a final counting result for the test image is obtained.

{\bf Ground Truth. }Our method uses the summation of a density map as ground truth. The summation of a density map is shown as follows.

\begin{equation}
t_n= \sum_{p\in Bn}D_{B_n}(p),
\label{eq:dens}
\end{equation}
where $t_n$ is ground truth of the $n$-th training patch, $B_n$ is the $n$-th training patch, $D_{B_n}$ is the density map of the $n$-th training patch. By using density map, our method slightly becomes the robust to vague target which existing on boundary of patches.

\section{Evaluation}
\label{sec:4}

\begin{figure}[t]
\centering
\includegraphics[width=0.95\linewidth]{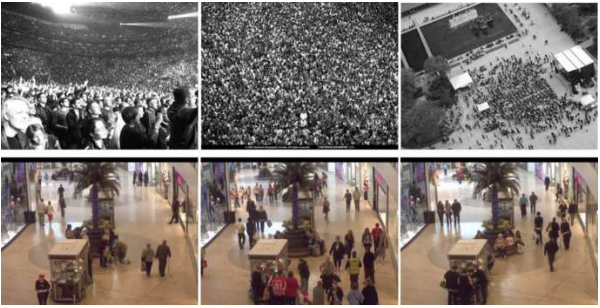}
\caption{Upper row is the examples of the UCF\_CC\_50 dataset. Lower row is the examples of the Mall dataset.}
\label{fig:dataset}
\end{figure}

Our MoC-CNN is evaluated using two challenging crowd counting datasets; the UCF\_CC\_50 dataset\cite{idrees2013multi} and the Mall dataset\cite{chen2012feature}. The examples of two datasets are shown in Figure \ref{fig:dataset}. Both datasets contain various appearance crowd caused by perspective and congestion.

In the rest of this section, we explain evaluation in each crowd counting dataset. In section \ref{ssec:3-1}, we show experimental results using the UCF\_CC\_50 dataset. Next, we show experimental result on the Mall dataset in section \ref{ssec:3-2}.

\subsection{UCF\_CC\_50 dataset}
\label{ssec:4-1}

Three images in Figure \ref{fig:dataset} on the upper row show the examples of the UCF\_CC\_50 dataset. The dataset contains 50 images. Each image contains people from 94 to 4543, the average number of crowd is 1280 persons. This dataset contains images filming at various scenes (e.g., demo, event and convention).

In this experiment, we use the same experimental setting as previous works\cite{pham2015count,zhang2015cross}. We randomly divide the dataset into 5 validation sets, and our method is evaluated by 5-fold cross-validation. The way to make training data refers to \cite{zhang2015cross}, and we randomly crop 1600 patches from a training image. Thus, the total number of training image is 64000 patches.

In this dataset, we use two evaluation metrics; Mean Absolute Error (MAE) and Mean Squares Deviation (MSD). Equations of these metrics are as follows.

\noindent\scalebox{.9}[1]{\box0}
\begin{eqnarray}
\varepsilon_{mae} = \frac{1}{N_{test}}\sum_{n=1}^{N_{test}}|t_n-y_n|,\\
\varepsilon_{msd} = \sqrt{\frac{1}{N_{test}}\sum_{n=1}^{N_{test}}(t_n-y_n)^2},
\label{eq:msd}
\end{eqnarray}
where $N_{test}$ is the total number of test image, $t_n$ is the ground truth of the $n$-th test image and $y_n$ is predicted value in the $n$-th test image.

\begin{figure}[t]
\begin{center}
\includegraphics[width=\linewidth]{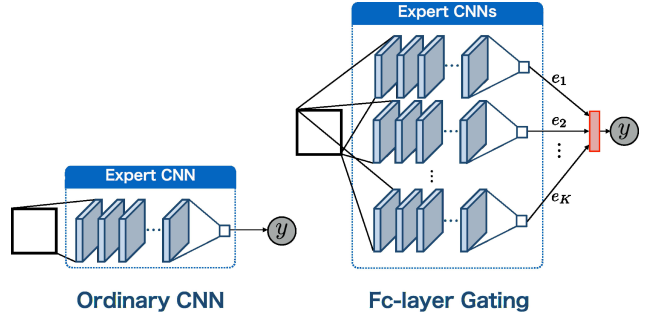}
\end{center}
\caption{The overview of comparative approaches. Left image is ordinary CNN which is equivalent to a expert CNN. Right image is Fc-layer gating. This method also combines expert CNN, but weights for expert CNN are fixed for all images.}
\label{fig:conv}
\end{figure}

To evaluate the effectiveness of the integration by gating CNN, we compare our method with two methods. The overview of those methods are shown in Figure \ref{fig:conv}. The first method is single ordinary CNN which is shown in left image of Figure \ref{fig:conv}. The architecture of the CNN is the same as our expert CNN. We call this method as Ordinary CNN.

The second method integrates expert CNNs using a fully-connected layer instead of using gating CNN. The overview of this method is shown in right image of Figure \ref{fig:conv}. This method also uses expert CNNs. The outputs of expert CNNs are fed into the fully-connected layer, and the fully-connected layer predicts the number of targets from the outputs of expert CNNs. Thus, the weight for integrating the CNNs is fixed for all test images while our method adaptively integrate expert CNNs. We call this method as Fc-layer Gating.

\begin{table}[t]
\begin{center}
\caption{Crowd counting result on the UCF\_CC\_50 dataset. }
\label{table:1}
\begin{tabular}{l c c }
\hline
\noalign{\smallskip}
 & MAE & MSD\\
\noalign{\smallskip}
\hline
\noalign{\smallskip}
Idrees \etal \cite{idrees2013multi} & 419.5 & - \\
Zhang \etal \cite{zhang2015cross} & 467.0 & 498.5 \\
Zhang \etal \cite{zhang2016single} & 377.6 & 509.1\\
O\~{n}oro \etal \cite{onoro2016towards} & 333.7 & 425.3\\
\noalign{\smallskip}
\hline
\noalign{\smallskip}
Ordinary CNN & 545.6 & 697.5\\
Fc-layer Gating & 509.6 & 670.0 \\
MoC-CNN & 361.7 & 493.3\\
\noalign{\smallskip}
\hline
\end{tabular}
\end{center}
\end{table}

\begin{figure*}[t]
\begin{center}
\includegraphics[width=\linewidth]{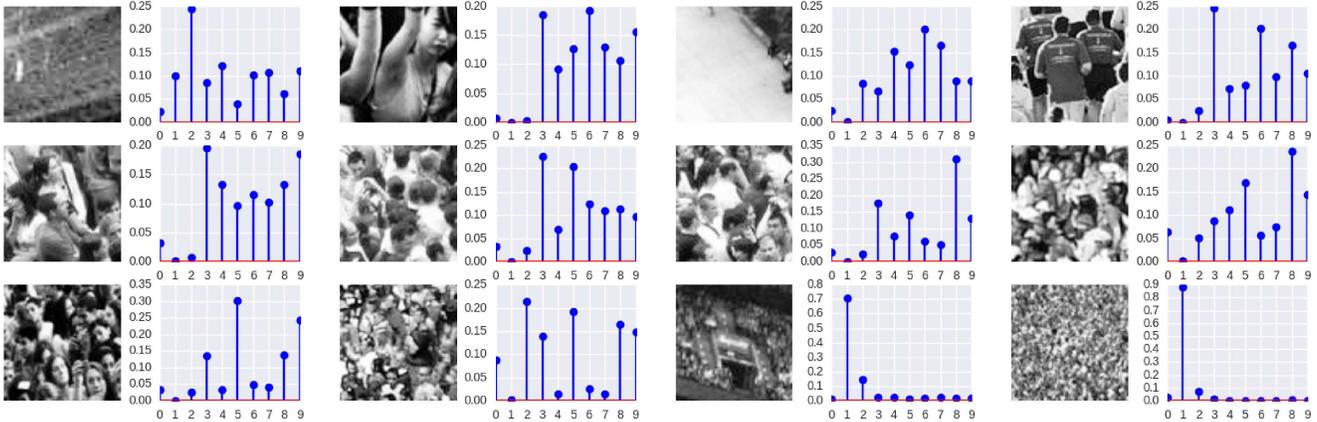}
\end{center}
\caption{Images in the UCF\_CC\_dataset and the outputs of gating CNN.}
\label{fig:ucf_result}
\end{figure*}

Experiment result is shown in Table \ref{table:1}. We compare MoC-CNN, above two compatative methods and five state-of-the-art methods. In the state-of-the-art method, Idrees \etal \cite{idrees2013multi} uses regression based method without CNN. Zhang \etal \cite{zhang2015cross} uses single CNN. Zhang \etal \cite{zhang2015cross} and O\~{n}oro \etal \cite{onoro2016towards} are CNN based methods, and those methods use CNN constructed by multiple feature extraction units. Walach \etal \cite{walach2016learning} uses multiple CNNs, and the first CNN predicts the number of targets. Then, the second CNN predicts the counting error of the first CNN, and the second CNN corrects the counting error of the first CNN.

MAE of the proposed method decreases 34\% and 29\% in comparison with the ordinary CNN and Fc-layer gating. We confirm the effectiveness of the integration by gating CNN. Moreover, We confirm that our method obtains comparable accuracy with the state-of-the-art methods.

O\~{n}oro \etal marked the best accuracy in Table \ref{table:1}. However, this method increases training data using data augmentation, and the number of training data of this method is larger than our method and other comparative methods. In methods using same experiment setting, MoC-CNN obtains the best accuracy.

Figure \ref{fig:ucf_result} shows test images and the outputs of gating CNN. In Figure \ref{fig:ucf_result}, images on upper row shows background and few humans, and images on lower row shows dense crowd. Graphs on right side of each image shows the outputs of gating CNN for each image. On the graphs, horizontal axis means each gating CNN, and vertical axis is the output of gating CNN. We confirm that the 6th and 7th outputs of gating CNN often react to few humans and background. On the other hand, the 2nd output of gating CNN often reacts the image of dense crowd, the 5th output of gating CNN reacts middle level density. Thus, gating CNN adaptively select expert CNNs according to the appearance of test images.

\subsection{Mall dataset}
\label{ssec:4-2}

Mall dataset contains 2000 frames. We use the same experimental setting as previous works \cite{chen2012feature}. We use the first 800 frames for training and the rest 1200 frames for evaluation. To make training data, we randomly crop 80 patches from a training image. The total number of training patches is 64000, it is the same as the experiment using the UCF\_CC\_50 dataset.

In this dataset, we use three metrics for evaluation; MAE, Mean Squared Error (MSE) and Mean Deviation Error (MDE). Equation of those metrics are shown as follows.

\noindent\scalebox{.9}[1]{\box0}
\begin{equation}
\varepsilon_{mse} = \frac{1}{N_{test}}\sum_{n=1}^{N_{test}}(t_n-y_n)^2,
\varepsilon_{mde} = \frac{1}{N_{test}}\sum_{n=1}^{N_{test}}\frac{|t_n-y_n|}{t_n},
\label{eq:msd}
\end{equation}

\begin{table}[t]
\begin{center}
\caption{Crowd counting result on the Mall dataset. }
\label{table:2}
\begin{tabular}{l | c c c}
\hline
\noalign{\smallskip}
 & MAE & MSE & MDE\\
\noalign{\smallskip}
\hline
\noalign{\smallskip}
LSSVR \cite{van2001automatic} & 3.51 & 18.2 & 0.108\\
KRR \cite{an2007face} & 3.51 & 18.1 & 0.108 \\
RFR \cite{liaw2002classification} & 3.91 & 21.5 & 0.121 \\
GPR \cite{chan2008privacy} & 3.72 & 20.1 & 0.115 \\
RR \cite{chen2012feature} & 3.59 & 19.0 & 0.105 \\
CA-RR \cite{chen2013cumulative} & 3.43 & 17.7 & 0.105\\
SSR \cite{loy2013semi} & - & 17.8 & -\\
CF \cite{pham2015count} & 2.50 & 10.0 & 0.080\\
Boosting CNN \cite{walach2016learning} & 2.01 & - & -\\
\hline
MoC-CNN & 2.75 & 13.4 & 0.087 \\
\noalign{\smallskip}
\hline
\end{tabular}
\end{center}
\end{table}

\begin{figure*}[t]
\begin{center}
\includegraphics[width=\linewidth]{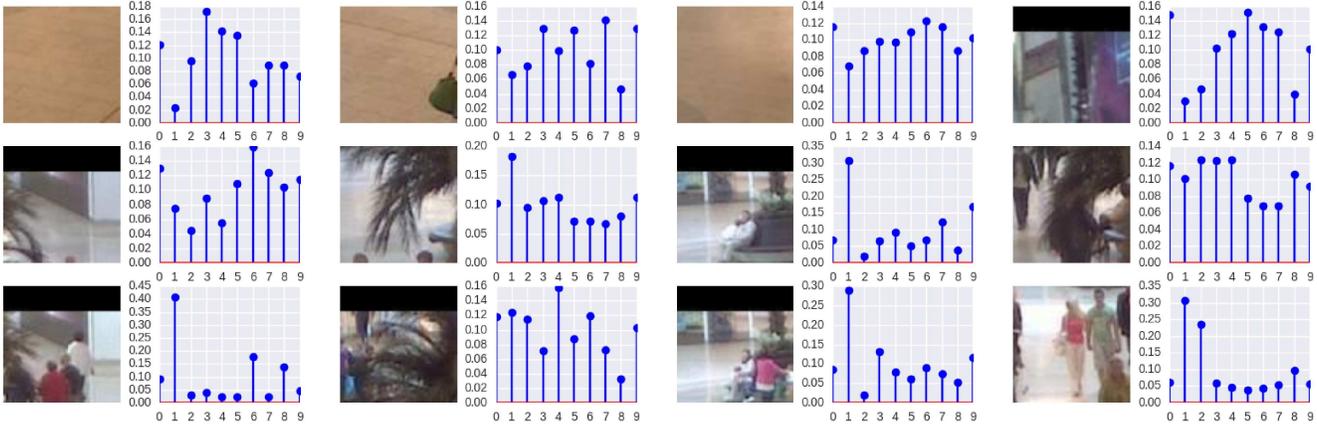}
\end{center}
\caption{Images in the Mall dataset and the outputs of gating CNN.}
\label{fig:mall_result}
\end{figure*}

In this experiment, we compare our method with the recent works; Least Square Support Vector Regression \cite{van2001automatic} (LSSVR), Kernel Ridge Regression\cite{an2007face} (KRR), Random Forest Regression \cite{an2007face} (RFR), Gaussian Process Regression \cite{chan2008privacy} (GPR), Multiple Output Ridge Regression \cite{chen2012feature} (RR), Cumulative Attribute Ridge Regression \cite{chen2013cumulative} (CA-RR), Semi-Supervised Regression \cite{loy2013semi} (SSR), Random Forest based method \cite{pham2015count} (CF) and multiple CNNs based method \cite{walach2016learning}(Boosting CNN).

Our method outperforms conventional methods except for CF and Boosting CNN. Although our method directly predicts the number of target. On the other hand, CF and Boosting CNN predicts density map, density map estimation has the robustness to vague target which existing on boundary of patches. Since training data of the proposed method contain a lot of vague targets, our method is affected by vague targets. Therefore, we consider that our method improves the accuracy by introducing density map.

Test images of the Mall dataset and the outputs of gating CNN are shown in Figure \ref{fig:mall_result}. Images on upper row are background image, and gating CNN predicts similar outputs to those background image. On the other hand, if small person exists in an image, the 2nd output of gating CNN strongly react. the variations of appearance on the Mall dataset is smaller than the UCF\_CC\_50 dataset. Thus, the output of gating CNN do not variously change as much as the result of the UCF\_CC\_50 dataset.


\section{Conclusion}
\label{sec:5}

We proposed MoC-CNN which integrates CNNs specialized to a specific appearance for crowd counting. We show the effectiveness of adaptive integration of some CNNs by the comparison with a CNN and integration using fixed weights. The proposed method obtains comparable result to conventional counting methods.

The proposed method is affected by vague training data. Therefore, we should use density map to improve the accuracy. Moreover, we may use the idea of Hierarchical Mixture of Experts. They are subjects for future works.


{\small
\bibliographystyle{ieee}

}

\end{document}